\documentclass[twoside,11pt]{article}

\usepackage{jmlr2e}
\usepackage{enumitem}
\usepackage[utf8]{inputenc}  
\usepackage{booktabs}        
\usepackage{pifont}          
\usepackage{xspace}
\usepackage{xcolor}
\usepackage{threeparttable}  

\newcommand{\skmultiflow}{\textsf{scikit-multiflow}\xspace}
\newcommand{\sklearn}{\textsf{scikit-learn}\xspace}
\newcommand{\weka}{\textsf{WEKA}\xspace}
\newcommand{\moa}{\textsf{MOA}\xspace}
\newcommand{\meka}{\textsf{MEKA}\xspace}
\newcommand{\streammodel}{\textsf{StreamModel}\xspace}
\newcommand{\streamevaluator}{\textsf{StreamEvaluator}\xspace}
\newcommand{\stream}{\textsf{Stream}\xspace}
\newcommand{\reffig}[1]{Figure~\ref{#1}}
\newcommand{\reftab}[1]{Table~\ref{#1}}

\newcommand{\cmark}{\ding{51}}  

\jmlrheading{1}{2018}{1-48}{4/25}{10/00}{Jacob Montiel and Jesse Read and Albert Bifet and Talel Abdessalem}

\ShortHeadings{
Scikit-Multiflow: A Multi-output Streaming Framework
}{Montiel and Read and Bifet and Abdessalem}
\firstpageno{1}

\begin{document}

\title{
Scikit-Multiflow: A Multi-output Streaming Framework
}

\author{\name Jacob Montiel \email jacob.montiel@telecom-paristech.fr \\
       \addr LTCI,
       Télécom ParisTech - Université Paris-Saclay\\
       Paris, FRANCE
       \AND
       \name Jesse Read \email jesse.read@polytechnique.edu \\
       \addr LIX, 
       École Polytechnique\\
       Palaiseau, FRANCE
       \AND
       \name Albert Bifet \email albert.bifet@telecom-paristech.fr \\
       \addr LTCI,
       Télécom ParisTech - Université Paris-Saclay\\
       Paris, FRANCE
       \AND
       \name Talel Abdessalem \email talel.abdessalem@enst.fr \\
       \addr LTCI, 
       Télécom ParisTech - Université Paris-Saclay\\
       Paris, FRANCE \\
       UMI CNRS IPAL \\
       National University of Singapore
       }

\editor{Leslie Pack Kaelbling}

\maketitle

\begin{abstract}
\skmultiflow is a multi-output/multi-label and stream data mining framework for the Python programming language. Conceived to serve as a platform to encourage democratization of stream learning research, it provides multiple state of the art methods for stream learning, stream generators and evaluators. \skmultiflow builds upon popular open source frameworks including \sklearn, \moa\ and \meka. Development follows the FOSS principles and quality is enforced by complying with PEP8 guidelines and using continuous integration and automatic testing. The source code is publicly available at \url{https://github.com/scikit-multiflow/scikit-multiflow}.
\end{abstract}


\section{Introduction}\label{sec:Introduction}
Recent years have witnessed the proliferation of \underline{F}ree and \underline{O}pen \underline{S}ource \underline{S}oftware (FOSS) in the research community. Specifically, in the field of Machine Learning, researchers have benefited from the availability of different frameworks that provide tools for faster development, allow replicability and \textbf{reproducibility} of results and foster collaboration. 

\textsf{scikit-learn}~\citep{sklearn} is the most popular open source software machine learning library for the Python programming language. It features various classification, regression and clustering algorithms including support vector machines, random forest, gradient boosting, k-means and DBSCAN, and is designed to inter-operate with the Python numerical and scientific packages NumPy and SciPy.

\moa~\citep{MOA} is the most popular open source framework for data stream mining, with a very active growing community. It includes a collection of machine learning algorithms (classification, regression, clustering, outlier detection, concept drift detection and recommender systems) and tools for evaluation. Related to the \weka project~\citep{WEKA}, \moa is also written in Java, while scaling to more demanding problems. 

The \meka project~\citep{MEKA} provides an open source implementation of methods for multi-label learning and evaluation. In multi-label classification, the aim is to predict multiple output variables for each input instance. This different from the `standard' case (binary, or multi-class classification) which involves only a single target variable.

Following the FOSS principles, we introduce \skmultiflow, a multi-output/multi-label and data stream framework for the Python programming language. \skmultiflow is inspired in the popular frameworks \sklearn, \moa and \meka.

\begin{table}[!b]
    \caption{Available methods. Methodologies on the left, and frameworks on the right of the vertical bar. }\label{table:methods}
    \begin{threeparttable}[t]
    \centering
    \scriptsize
    \begin{tabular}{@{}lcccccc|ccccc@{}}
    \toprule
                            &                                         &                                       &                                        &                                      &                                       &                                          & \multicolumn{2}{c}{Java}                                              & \multicolumn{2}{c}{Python}   \\ 
                            \cmidrule(lr){8-9} \cmidrule(lr){10-11}
                            & \rotatebox{90}{\textbf{Classification}} & \rotatebox{90}{\textbf{Regression}}   & \rotatebox{90}{\textbf{Single-Output}} & \rotatebox{90}{\textbf{Multi-Label}} & \rotatebox{90}{\textbf{Multi-Output}} & \rotatebox{90}{\textbf{Drift Detection}} & \rotatebox{90}{\textbf{MOA}}    & \rotatebox{90}{\textbf{MEKA}\tnote{$\dagger$}} & \rotatebox{90}{\textbf{scikit-learn}\tnote{$\dagger$}}  & \rotatebox{90}{\textbf{scikit-multiflow}}  & \textbf{Reference} \\ \midrule
    kNN                     & \cmark                                  &                                       & \cmark                                 & \cmark                               &                                       &                                          & \cmark                          & \cmark                                         & \cmark                                                  & \cmark                                     & \cite{bishop2006pattern}  \\
    kNN + ADWIN             & \cmark                                  &                                       & \cmark                                 & \cmark                               &                                       &                                          & \cmark                          &                                                &                                                         & \cmark                                     & \cite{Bifet-et-al-2018}  \\
    SAM kNN                 & \cmark                                  &                                       & \cmark                                 & \cmark                               &                                       & \cmark                                   & \cmark                          &                                                &                                                         & \cmark                                     & \cite{Losing2017} \\
    Hoeffding Tree          & \cmark                                  &                                       & \cmark                                 & \cmark                               &                                       &                                          & \cmark                          &                                                &                                                         & \cmark                                     & \cite{Domingos2000}  \\
    Hoeffding Adaptive Tree & \cmark                                  &                                       & \cmark                                 & \cmark                               &                                       & \cmark                                   & \cmark                          &                                                &                                                         & \cmark                                     & \cite{Bifet-et-al-2018}  \\
    Adaptive Random Forest  & \cmark                                  &                                       & \cmark                                 & \cmark                               &                                       & \cmark                                   & \cmark                          &                                                &                                                         & \cmark                                     & \cite{Gomes2017}  \\
    Oza Bagging             & \cmark                                  &                                       & \cmark                                 & \cmark                               &                                       & \cmark                                   & \cmark                          &                                                &                                                         & \cmark                                     & \cite{Oza2005}  \\
    Leverage Bagging        & \cmark                                  &                                       & \cmark                                 & \cmark                               &                                       & \cmark                                   & \cmark                          &                                                &                                                         & \cmark                                     & \cite{Bifet-et-al-2018} \\
    Multi-output Learner    & \cmark                                  & \cmark                                & \cmark                                 & \cmark                               & \cmark                                & \tnote{*}                                & \cmark                          & \cmark                                         & \cmark                                                  & \cmark                                     & \cite{bishop2006pattern}  \\
    SGD                     & \cmark                                  & \cmark                                & \cmark                                 & \cmark                               &                                       &                                          & \cmark                          & \cmark                                         & \cmark                                                  & \cmark                                     & \cite{bishop2006pattern}  \\
    Naive Bayes             & \cmark                                  &                                       & \cmark                                 & \cmark                               &                                       &                                          & \cmark                          & \cmark                                         & \cmark                                                  & \cmark                                     & \cite{bishop2006pattern}  \\
    MLP                     & \cmark                                  & \cmark                                & \cmark                                 & \cmark                               &                                       &                                          &                                 &   \cmark                                       & \cmark                                                  & \cmark                                     & \cite{bishop2006pattern}  \\
    ADWIN                   &                                         &                                       &                                        &                                      &                                       & \cmark                                   & \cmark                          &                                                &                                                         & \cmark                                     & \cite{Bifet-et-al-2018} \\
    DDM                     &                                         &                                       &                                        &                                      &                                       & \cmark                                   & \cmark                          &                                                &                                                         & \cmark                                     & \cite{Gama2004}  \\
    EDDM                    &                                         &                                       &                                        &                                      &                                       & \cmark                                   & \cmark                          &                                                &                                                         & \cmark                                     & \cite{Bifet-et-al-2018}  \\
    Page Hinkley            &                                         &                                       &                                        &                                      &                                       & \cmark                                   & \cmark                          &                                                &                                                         & \cmark                                     & \cite{Page1954} \\
    \bottomrule
    \end{tabular}
    \begin{tablenotes}
     \item[*] Depending on the base learner.
     \item[$\dagger$] We have only listed incremental methods for data-streams; MEKA and scikit-learn have many other batch-learning models available. MEKA in particular, has many problem-transformation methods which may be incremental depending on the base learner (it is able to use those from the MOA framework).
   \end{tablenotes}
    \end{threeparttable}
\end{table}

As a multi-output streaming framework, \skmultiflow serves as a bridge between research communities that have flourished around the aforementioned popular frameworks, providing a common ground where they can thrive. \skmultiflow assists on the democratization of Stream Learning by bringing this research field closer to the Machine Learning community, given the increasing popularity of the Python programing language. The objective is two-folded: First, fill the void for a stream learning framework in Python, which can interact with available tools such as \sklearn and extends the set of available state-of-the-art methods on this platform. Second, provide a set of tools to facilitate the development of stream learning research.

It is important to notice that \skmultiflow complements \sklearn, whose primary focus is batch learning, expanding the set of free and open source tools for Stream Learning. In addition, \skmultiflow can be used within Jupyter Notebooks, a popular interface in the Data Science community. Special focus in the design of \skmultiflow is to make it friendly to new users and familiar to experienced ones.

\skmultiflow contains stream generators, learners, change detectors and evaluation methods. Stream generators include: Multi label, Random-RBF, Random-RBF with drift, Random Tree Regression, SEA and Waveform. Learners and change detectors are listed in \reftab{table:methods}. Available evaluators are prequential and hold-out.

\section{Notation and background}\label{sec:Notation}
Consider a continuous stream of data $A=\{(\vec{x}_t,y_t)\} | t = 1,\ldots,T$ where $T \rightarrow \infty$. $\vec{x}_t$ is a feature vector and $y_t$ the corresponding target where $y$ is continuous in the case of regression and discrete for classification. The objective is to predict the target $y$ for an unknown $\vec{x}$. Two classes are considered in \textit{binary} classification, $y\in \{0,1\}$, while $K>2$ labels are used in \textit{multi-label} classification, $y\in \{1,\ldots,K\}$. For both \textit{binary} and \textit{multi-label} classification only one class is assigned per instance. On the other hand, in multi-output learning $y$ is a targets vector and $\vec{x}_i$ can be assigned multiple-targets at the same time.

Different to batch learning, where all data is available for training $train(X, y)$; in stream learning, training is performed incrementally as new data is available $train(\vec{x}_i, y_i)$. Performance $P$ of a given model is measured according to some loss function that evaluates the difference between the set of expected labels $Y$ and the predicted ones $\hat{Y}$. Hold-out evaluation is a popular performance evaluation method for batch and stream settings, where tests are performed in a separate test set. Prequential-evaluation or interleaved-test-then-train evaluation, is a popular performance evaluation method for the stream setting only, where tests are performed on new data before using it to train the model.

\begin{figure}[!ht]
	\centering
    \includegraphics[width=0.9\textwidth]{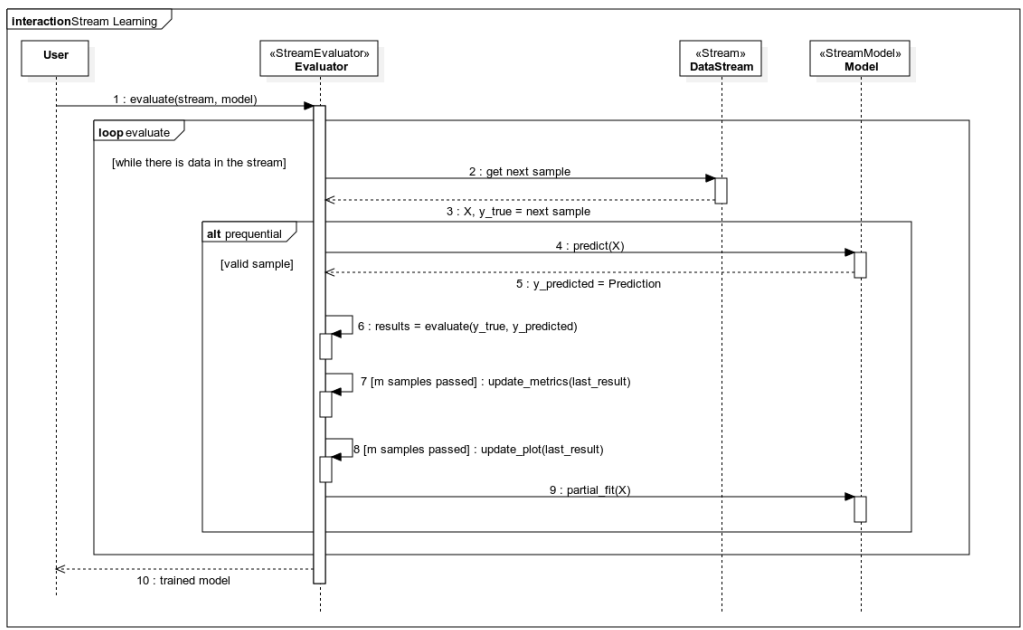}
    \caption{Training and testing a stream model using \skmultiflow. Sequence corresponds to prequential evaluation.}\label{fig:learn_seq}
\end{figure}

\section{Architecture}\label{sec:Architecture}
The \streammodel class is the base class in \skmultiflow. It contains the following abstract methods:
\begin{itemize}[noitemsep]
\item \textsf{fit} --- Trains a model in a batch fashion. Works as a an interface to batch methods that implement a \textsf{fit()} functions such as \sklearn methods.
    \item \textsf{partial\_fit} --- Incrementally trains a stream model.
    \item \textsf{predict} --- Predicts the target's value in supervised learning methods.
    \item \textsf{predict\_proba} --- Calculates the probability of a sample pertaining to a given class in classification problems.
\end{itemize}

An \streammodel object interacts with two other objects: an \stream object and (optionally) an \streamevaluator object. The \stream object provides a continuous flow of data on request. The \streamevaluator performs multiple tasks: query the stream for data, train and test the model on the incoming data and continuously tracks the model's performance.  

The sequence to train a Stream Model and track its performance using prequential evaluation in \skmultiflow is outlined in \reffig{fig:learn_seq}. 

\section{Development}\label{sec:Development}
\skmultiflow~is distributed under the BSD License. Development follows the FOSS principles and includes:
\begin{itemize}[noitemsep]
\item A webpage including documentation: \url{https://scikit-multiflow.github.io/}.
\item A web platform for users: \url{https://goo.gl/AyPsMj}
\item Version control via git. The source code is publicly available at\\ \url{https://github.com/scikit-multiflow/scikit-multiflow}
\item Package deployment and software quality are enforced via continuous integration and automatic testing, \url{https://travis-ci.org/scikit-multiflow/scikit-multiflow}
\item A user guide: \url{https://scikit-multiflow.github.io/scikit-multiflow/user-guide}
\end{itemize}



\bibliography{biblio}

\end{document}